\documentclass[a4paper]{article}
\usepackage{color,xcolor}
\usepackage{INTERSPEECH_v2}
\usepackage[font=small,skip=2pt]{caption}
\usepackage{balance}
\usepackage{microtype}
\usepackage{float}
\usepackage{cite}
\usepackage[disable]{todonotes}
\newcommand{\calS}{\mathcal{S}}
\newcommand{\calX}{\mathcal{X}}
\newcommand{\R}{\mathbb{R}}

\title{Query-by-Example Search \\ with Discriminative Neural Acoustic Word Embeddings}
\name{Shane Settle$^1$, Keith Levin$^2$, Herman Kamper$^1$, Karen Livescu$^1$}
\address{$^1$TTI-Chicago, USA $^2$Johns Hopkins University, USA}
\email{\{settle.shane, kamperh, klivescu\}@ttic.edu, klevin@jhu.edu}

\begin{document}
\maketitle

\begin{abstract}
Query-by-example search often uses dynamic time warping (DTW) for comparing queries and proposed matching segments.  Recent work has shown that comparing speech segments by representing them as fixed-dimensional vectors --- acoustic word embeddings --- and measuring their vector distance (e.g., cosine distance) can discriminate between words more accurately than DTW-based approaches. We consider an approach to query-by-example search that embeds both the query and database segments according to a neural model, followed by nearest-neighbor search to find the matching segments.  Earlier work on embedding-based query-by-example, using template-based acoustic word embeddings, achieved competitive performance. We find that our embeddings, based on recurrent neural networks trained to optimize word discrimination, achieve substantial improvements in performance and run-time efficiency over the previous approaches.
\end{abstract}
\noindent\textbf{Index Terms}: query-by-example, acoustic word embeddings, word discrimination, recurrent neural networks

\section{Introduction}
Query-by-example speech search (QbE) is the task of searching for a spoken query term (a word or phrase) in a collection of speech recordings.  Unlike keyword search and spoken term detection, where the search terms are given as text, QbE involves matching audio segments directly. This task arises naturally when the search terms may be out-of-vocabulary~\cite{shen2009comparison,parada2009query}, in hands-free settings, or in low- or zero-resource settings~\cite{szoke2015query}.

For QbE in high-resource settings, one can train a model to map the audio query to a sequence of subword units, such as phonemes, and search for this sequence in a lattice built from the search collection~\cite{allauzen2004general,parada2009query}. This approach requires very significant resources, since it involves much the same process as training a full speech recognition system.

In low-resource settings, typical approaches for this task use dynamic time warping (DTW) to determine the similarity between audio segments. Early approaches to low-resource QbE were based on performing DTW alignment of the query against a search collection either exactly~\cite{HazSheWhi2009,zhang2009unsupervised} or approximately~\cite{zhang2011piecewise,JanVan2012,ManAng2013}.

\begin{figure}[t]
\centering
\includegraphics[width=\linewidth]{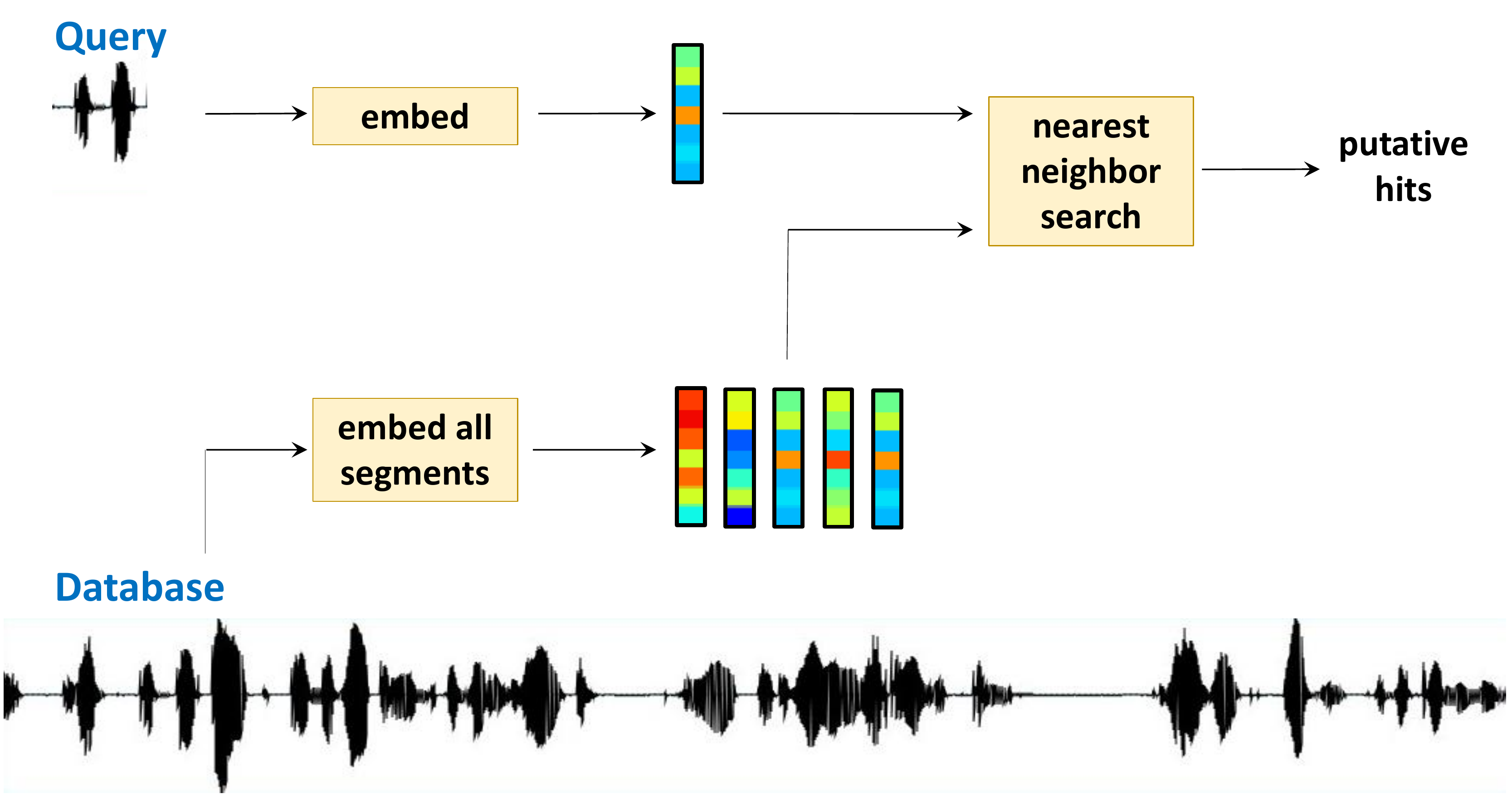}
\caption{Embedding-based query-by-example search.}
\label{fig:segmental_qbe}
\end{figure}

An alternative to DTW for QbE, which we explore in this paper, is to represent variable-duration speech segments as fixed-dimensional vectors and directly measure similarity between them via a simple vector distance. In this approach, shown in Figure~\ref{fig:segmental_qbe}, the query is embedded using an \textit{acoustic word embedding} function, producing a vector representation of the query. All potential segments in the search collection are then represented as vectors using the same embedding function. The putative hits (matches) correspond to those segments in the search collection that are closest to the query in the fixed-dimensional embedding space. This type of approach requires preprocessing steps for learning the embedding function and generating the embeddings for the search collection. At test time, efficient approximate nearest-neighbor search can greatly speed up computation.

In prior work, Levin {\it et al.}~\cite{LevJanVan2015} used a template-based acoustic word embedding function, and showed that this type of embedding-based QbE search can greatly speed up search compared to a purely DTW-based system, while matching or improving performance. Their template-based embedding approach does not require any labeled supervision. However, in many practical settings, a limited amount of training data might be available. In this work we consider this low-resource setting; in particular, we use acoustic word embeddings based on neural models learned to discriminate between words given a limited (roughly 2-hour) training set.

We build on a growing body of work on neural network-based acoustic word embeddings~\cite{chung2016unsupervised,kamper2016deep,settle2016discriminative,he2017multi,AudETAL2017}. In several of these studies, neural approaches are shown to far outperform template-based embeddings (such as those used in~\cite{LevJanVan2015}) on an isolated-word discrimination task, which can be viewed as a proxy for QbE. Here, we use the neural embedding approach of~\cite{settle2016discriminative}, based on Siamese recurrent neural networks, and incorporate these into a complete QbE system using the embedding-based approach of Levin {\it et al.}~\cite{LevJanVan2015}. We show that these neural embeddings, trained only on a small amount of labeled data, achieve large improvements in true QbE performance.

\section{Neural embedding-based QbE}

As illustrated in Figure~\ref{fig:segmental_qbe}, embedding-based query-by-example (QbE) consists of an embedding method and a nearest neighbor search component. We first describe our neural acoustic word embedding approach, and then give details of the embedding-based QbE search system in which the embeddings are used.

\begin{figure}
\centering
\includegraphics[width=\linewidth]{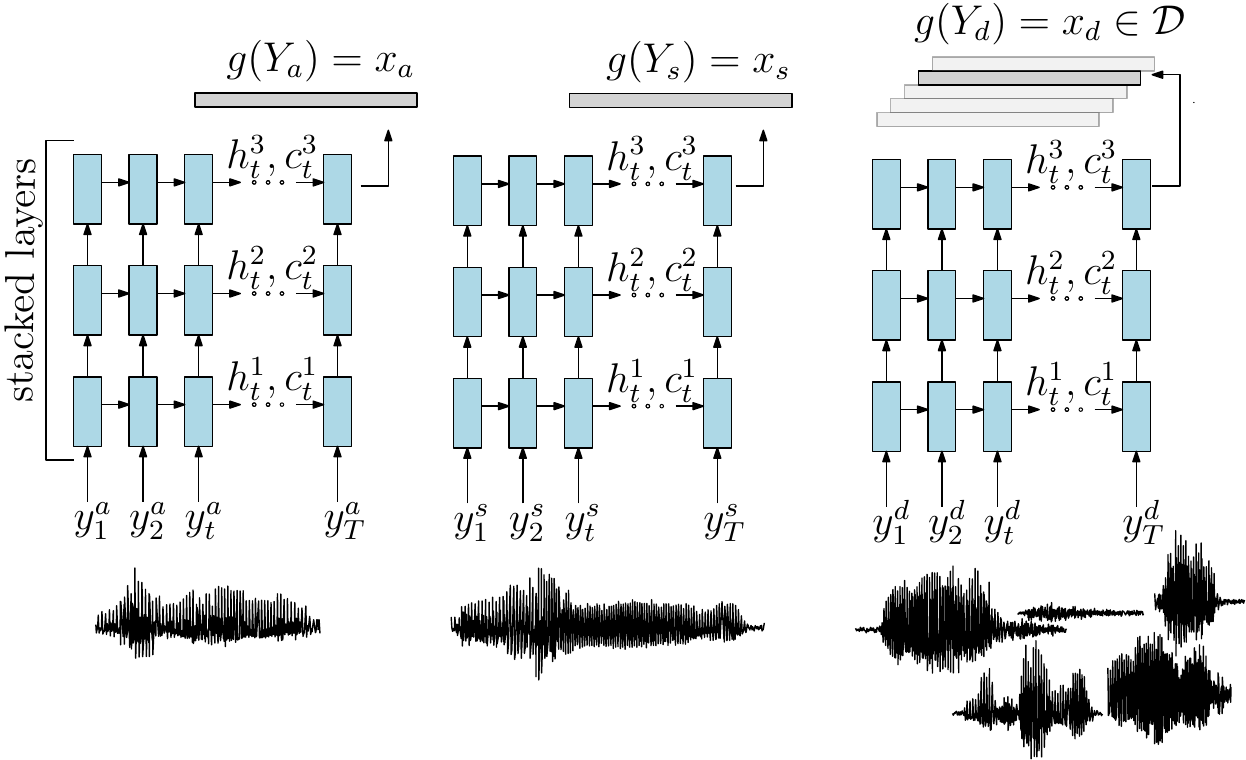}	
\caption{Triplet Siamese training setup (unidirectional LSTM depicted for simplicity).}
\label{fig:siamese}
\end{figure}

\subsection{Neural acoustic word embeddings (NAWEs)}
\label{ssec:nawes}

An acoustic word embedding function $g$ maps a variable-length speech segment $Y=y_1, y_2, \ldots, y_T$, where each $y_i$ is an acoustic feature frame, to a single embedding vector $x\in\mathbb{R}^d$. Ideally, $g$ should map instances of the same word to nearby vectors, while different words are mapped far apart. Once segments are embedded, they can be compared by computing a vector distance between their embeddings, rather than using DTW.

In~\cite{LevHenJanLiv2013}, Levin {\it et al.} proposed a template-based approach for the embedding function $g$.  For a target segment, a {\it reference vector} is defined as the vector of DTW alignment costs to a set of template segments. Dimensionality reduction (based on Laplacian eigenmaps~\cite{belkin+niyogi_neurocomp03}) is then applied to the reference vector to obtain an embedding in $\mathbb{R}^d$. This template-based embedding approach was subsequently used for unsupervised speech recognition in~\cite{kamper+etal_taslp16} and, more importantly, the full QbE system of~\cite{LevJanVan2015}, which we consider to be a baseline in our experiments.

Recently, neural acoustic word embeddings (NAWEs) have been proposed as an alternative~\cite{chung2016unsupervised,kamper2016deep,settle2016discriminative,he2017multi,AudETAL2017}. In this recent work, NAWEs have achieved much better performance than the template-based approach, but only in an isolated-word discrimination task that can be seen as a proxy for QbE~\cite{carlin+etal_icassp11}. Here, we specifically focus on the NAWE approach developed in~\cite{settle2016discriminative}, where it was shown that embeddings based on Long Short-Term Memory (LSTM)~\cite{LSTM} networks outperform competing feedforward and convolutional methods. Rather than the proxy task, here we apply these NAWEs in a complete QbE system.

Concretely, we use the concatenation of the hidden representations from a deep bidirectional LSTM network as our embedding function, i.e.\ $x = g(Y) = [\overrightarrow{h_T}; \overleftarrow{h_1}]$, where $\overrightarrow{h_T}, \overleftarrow{h_1}$ refer to the final hidden state vector from the forward and backward LSTMs, respectively. This LSTM is trained using a Siamese weight-sharing scheme~\cite{bromley+etal_ijpr93} depicted in Figure~\ref{fig:siamese} with a contrastive triplet loss~\cite{chopra2005learning,socher2014grounded}, $l_{\textrm{cos hinge}}(Y_a, Y_s)$, defined as
\vspace{-0.15cm}
\begin{equation*}
  \max \left\{0, m + d_{\cos}(x_a,x_s) -  \max_{x_d \in \mathcal{D}} d_{\cos}(x_a,x_d) \right\}
\end{equation*}
\vspace{-0.05cm}
In this definition, $Y_a$ and $Y_s$ are two segments that have the same word label, and $x_a, x_s$ are their embeddings as output by the neural embedding network. The goal is to push the embeddings $x_a$ and $x_s$ together, until they are closer to each other by a margin $m$ than the embedding $x_d$ of a negative example. Here $d_{\cos}(x_1, x_2) = (1 - \cos(x_1, x_2))$ is the cosine distance between vectors $x_1$ and $x_2$. Rather than sampling a single negative example as in~\cite{kamper2016deep}, or keeping track of confusion statistics as in~\cite{settle2016discriminative}, we sample a set of $k$ embedded segments $\mathcal{D}$ from the whole training set with labels different from $Y_a$ and consider only the example embedding, $x_d \in \mathcal{D}$, that most violates the margin constraint. This improves both performance and rate of convergence on the proxy task.

\subsection{Embedding-based QbE}
Our system needs to quickly retrieve from a large collection those segments nearest to a given spoken query. For this, we use the Segmental Randomized Acoustic Indexing and Logarithmic-Time Search (S-RAILS) system \cite{LevJanVan2015}, an embedding-based QbE approach. Although S-RAILS was first applied using the template-based embedding method, it is agnostic to the embedding type, and here we apply it to our neural embeddings.

S-RAILS provides a simple platform for performing approximate nearest neighbor search over vectors, relying on a version of locality-sensitive hashing (LSH)~\cite{IndMot1998,Charikar2002}. Let $\calX = \{x_1,x_2,\dots,x_N\} \in \R^d$ be the search collection. LSH is a method for representing vectors in $\R^d$ as bit vectors, referred to as signatures, such that if two vectors $x_i,x_j \in \calX$ are close under the cosine distance, then their signatures $s_i,s_j \in \{0,1\}^b$ will agree in most of their entries.

S-RAILS uses LSH to replace the comparatively expensive $d$-dimensional cosine distance between embeddings with a fast approximation. S-RAILS arranges the signatures $s(\calX) = \{ s_i : 1 \le i \le N \}$ into a lexicographically sorted list $\calS$. Given a query vector $q \in \R^d$, we map $q$ to its LSH signature $s = s(q) \in \{0,1\}^b$, and find its location in the sorted signature list $\calS$ in $O(\log b)$ time. A set of (approximate) near neighbors to $q$ can be read off this list by looking at the $B$ entries appearing before $s$ and the $B$ entries after $s$. Bits appearing earlier in the signature have far more influence on whether or not two vectors $x_i,x_j \in \calX$ will be judged similar.  To ameliorate this effect, S-RAILS performs this lexicographic lookup under $P$ different permutations of the bits: $\pi_1,\pi_2,\dots,\pi_P \in S_b$.

S-RAILS has three parameters: the signature length $b$, the {beamwidth} $B$, and the number of permutations $P$. Increasing any of these parameters will tend to improve performance either because it increases the fidelity of our approximation to the cosine distance (in the case of $b$ and $P$) or because it improves recall (in the case of $B$). However, any such improvements come at the cost of increased memory required to store the index and the permuted lists (in the case of $b$ and $P$) and increased runtime (in the case of $B$ and, to a lesser extent, $b$ and $P$). All told, building the index requires $O( P b N \log N )$ time in the worst case, and querying the index requires $O( B + P b \log N )$ time.

\section{Experimental setup}
We use data from the Switchboard corpus of (primarily American) English conversational telephone speech~\cite{Switchboard}. For training the NAWE model, we use a training set consisting of approximately 10k word segments covering less than 2 hours of speech taken from conversation sides distinct from those used to extract the query set and the evaluation collection. The size of this set is comparable to those used for training in prior work on acoustic word embeddings~\cite{kamper+etal_icassp15,jansen+etal_icassp13b,settle2016discriminative}. As acoustic features, we use 39-dimensional MFCC+$\Delta$+$\Delta\Delta$s. For QbE, we partition Switchboard into a 37-hour set from which to draw our query terms, a 48-hour development search collection on which to tune parameters of S-RAILS, and a 433-hour evaluation set. These partitions are identical to those used in prior work~\cite{JanVan2012,LevJanVan2015} for the QbE task. We use a set of 43 query words previously used in~\cite{JanVan2012,LevJanVan2015}, which were chosen subject to the constraints that the median word duration of each type across the entire corpus is at least 0.5 seconds and the orthographic representation of each word type has at least six characters~\cite{JanVan2012,LevHenJanLiv2013,LevJanVan2015}. Each word type appears 20 to 162 times in the query set, 2 to 188 times in the development search collection, and 39 to 1386 times in the evaluation set.

For our NAWE model (see  Section~\ref{ssec:nawes}), we use a stacked 3-layer bidirectional LSTM with 256 hidden units in each direction; the embeddings produced by the model are therefore 512-dimensional. Dropout is applied with probability $0.3$ between LSTM layers. For the margin of the contrastive loss, $l_{\textrm{cos\ hinge}}$, we use $m = 0.5$, and we sample $k = 10$ negative instances per anchor segment. We use the Adam optimization algorithm~\cite{kingma2014adam} with a batch size of 32, learning rate of $0.001$, $\beta_1 = 0.9$, $\beta_2 = 0.999$, and $\epsilon=1\cdot10^{-8}$. We tuned these parameters based on development set performance on the isolated word discrimination task of~\cite{carlin+etal_icassp11}. For our QbE evaluation experiments, we trained all models for $100$ epochs.

We evaluated the quality of search results according to three commonly used metrics: \emph{figure-of-merit} (FOM), \emph{oracular term weighted value} (OTWV), and \emph{precision at 10} (P@10). FOM is the recall averaged over the ten operating points at which the false alarm rate per hour of search audio is equal to $1,2,\dots,10$. OTWV is a query-specific weighted difference between the recall and the false alarm rate (further explanation can be found in~\cite{MillerEtAl2007}). P@10 is the fraction of the ten top-scoring results that are correct matches to the query.

Since the multiple query examples within each query type can have significant variation, we report {\it average median example} and {\it average maximum example} scores for each of these three metrics.  That is, we compute the median and maximum score over all examples of each query type, and report an unweighted arithmetic mean across the 43 query types.

\begin{figure}[t]
  \centering
  \includegraphics[width=\linewidth]{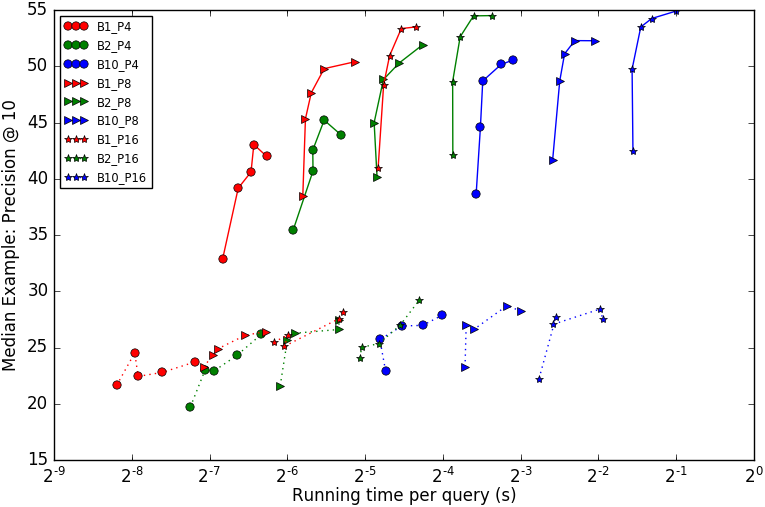}
  \caption{Performance (median P@10) of S-RAILS (dotted) and S-RAILS+NAWE (solid) on the development search collection.  Each sequence of connected points indicates results for a fixed permutation number (P) and beamwidth (B) while signature length (b) is varied from 128 to 2048.  In the legend, ``B$x$\_P$y$'' indicates a beamwidth of $1000x$ and number of permutations $y$.}  
  \label{fig:Pat10}
\end{figure}

\section{Results}

We first present QbE performance on the development data in order to show how performance differs across parameter settings, and then give evaluation results. In this section, we refer to the QbE system that employs the original template-based embeddings simply as S-RAILS, and to the system with neural embeddings as S-RAILS+NAWE.

\begin{table}[t]
\centering
\renewcommand{\arraystretch}{1.2}

  \caption{Effect of signature length $b$ on S-RAILS+NAWE performance
	on the development set, for $P=16$ permutations
	and beamwidth $B=2,000$.}
  \label{tab:nn3:siglensweep}
\begingroup \eightpt
  \begin{tabular}{@{} r | c | c | c | c | c | c | }
	\cline{2-7}
		 & \multicolumn{3}{|c|}{ \textbf{Median Example} }
		& \multicolumn{3}{|c|}{ \textbf{Best Example} } \\
		\hline
		$b$ & \textbf{FOM} & \textbf{OTWV} & \textbf{P@10}
		& \textbf{FOM} & \textbf{OTWV} & \textbf{P@10} \\
		\hline
	128  & 62.1 & 37.4 & 42.1 & 81.7 & 60.8 & 83.8\\
	256  & 67.2 & 42.6 & 48.6 & 83.0 & 65.4 & 84.9\\
	512  & 68.2 & 44.8 & 52.6 & 83.6 & 65.9 & 84.9\\
	1024 & 69.1 & 46.5 & 54.5 & 84.1 & 66.7 & 84.8\\
	2048 & 70.4 & 48.3 & 54.5 & 85.0 & 66.8 & 86.0\\
	\hline
  \end{tabular}

\endgroup

  \vspace{10pt}

  \caption{Effect of number of permutations $P$ on S-RAILS+NAWE performance on the
	development set, for signature length $b=1024$ and beamwidth $B=2,000$.}
  \label{tab:nn3:permsweep}
\begingroup \eightpt
  \begin{tabular}{ r | c | c | c | c | c | c | }
	\cline{2-7}
		 & \multicolumn{3}{|c|}{ \textbf{Median Example} }
		& \multicolumn{3}{|c|}{ \textbf{Best Example} } \\
		\hline
		$P$ & \textbf{FOM} & \textbf{OTWV} & \textbf{P@10}
		& \textbf{FOM} & \textbf{OTWV} & \textbf{P@10} \\
		\hline
	4  & 48.8 & 33.2 & 45.2 & 75.2 & 59.0 & 83.0\\
	8  & 60.9 & 41.0 & 50.3 & 80.3 & 63.8 & 85.0\\
	16 & 69.1 & 46.5 & 54.5 & 84.1 & 66.7 & 84.8\\
	\hline
  \end{tabular}
\endgroup

   \vspace{10pt}

  \caption{Effect of beamwidth $B$ on S-RAILS+NAWE performance
	on the development set,
	for signature length $b=1024$ and $P=16$ permutations. }
	\label{tab:nn3:bwsweep}
  \centering
\begingroup \eightpt
  \begin{tabular}{@{} r | c | c | c | c | c | c | }
	\cline{2-7}
		 & \multicolumn{3}{|c|}{ \textbf{Median Example} }
		& \multicolumn{3}{|c|}{ \textbf{Best Example} } \\
		\hline
		$B$ & \textbf{FOM} & \textbf{OTWV} & \textbf{P@10}
		& \textbf{FOM} & \textbf{OTWV} & \textbf{P@10} \\
		\hline
	1000  & 65.8 & 44.8 & 53.4 & 83.0 & 65.6 & 85.0\\
	2000  & 69.1 & 46.5 & 54.5 & 84.1 & 66.7 & 84.8\\
	10000 & 74.6 & 49.5 & 54.2 & 86.3 & 67.9 & 84.8\\
	\hline
  \end{tabular}
\endgroup
\end{table}

\begin{figure*}[t]
  \centering
  \includegraphics[width=\linewidth]{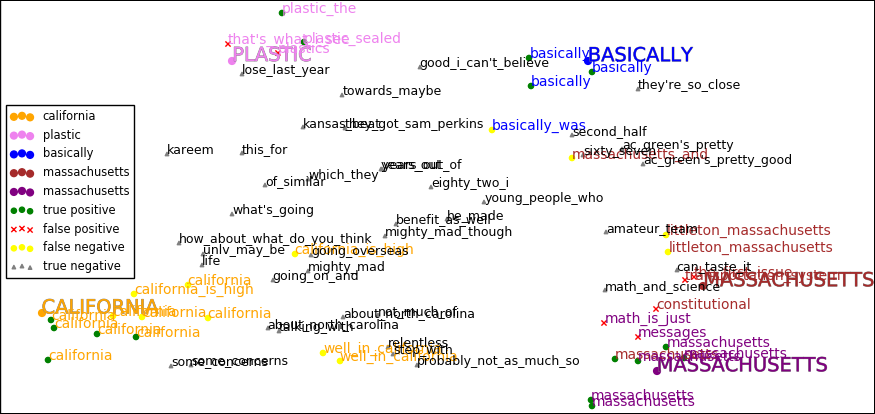}
  \caption{Embeddings of queries and their top hits, visualized in two dimensions using t-SNE~\cite{maaten2008visualizing}.  Queries are shown in capital letters.  The top several hits for each query are shown in the same color as the query. Random segments from the search collection and their associated transcriptions are shown in gray.}
  \label{fig:queries_and_top_hits}
\end{figure*}

\begin{table*}
  \caption{ Comparison of QbE system performance on the evaluation set. }
  \vspace{-0.25cm}
  \label{tab:eval}
  \begin{center}
  \begin{tabular}{ r | r | r | r | r | r | r | r }
	\cline{2-7}
		 & \multicolumn{3}{|c|}{ \textbf{Median Example} }
		& \multicolumn{3}{|c|}{ \textbf{Best Example} } \\
		\hline
		System & \textbf{FOM} & \textbf{OTWV} & \textbf{P@10}
		& \textbf{FOM} & \textbf{OTWV} & \textbf{P@10}
	& \textbf{Query time (s)} \\
		\hline
		RAILS \cite{JanVan2012}     &  6.7 &  2.7 & 44.0 & 20.7 & 10.4 & 84.4 & 24.7 \\
		S-RAILS (baseline) & 24.5 & 14.4 & 34.5 & 46.2 & 26.6 & 87.4 & 0.078 \\
		S-RAILS+NAWE (ours)  & 43.3 & 22.4 & 60.2 & 65.4 & 43.3 & 95.1 & 0.38 \\
		\hline
  \end{tabular}
  \end{center}
\end{table*}

\subsection{Development set performance}
Figure~\ref{fig:Pat10} shows development set performance, in terms of median P@10, for the baseline QbE system using template-based embeddings (S-RAILS) and our system using NAWEs (S-RAILS+NAWE). Tables~\ref{tab:nn3:siglensweep},~\ref{tab:nn3:permsweep}, and~\ref{tab:nn3:bwsweep} show development set performance for S-RAILS+NAWE as the signature length $b$, permutations $P$, and beamwidth $B$ are varied, respectively.

Figure~\ref{fig:Pat10} shows that neural embeddings improve the performance of S-RAILS by large margins at all running time operating points.  This figure also shows that increased signature length yields much larger improvements in P@10 for S-RAILS+NAWE than it does for the baseline S-RAILS system. Significant improvements in P@10 can be seen when holding fixed any combination of settings for $P$ and $B$.  Our performance on P@10 saturates with signatures around 1024 bits, while S-RAILS' saturates, for the most part, at 256 bits.

Again in contrast to the S-RAILS system, our method responds strongly to increases in the number of permutations used. In both Figure~\ref{fig:Pat10} and Table~\ref{tab:nn3:permsweep}, adjustment to this parameter improves performance consistently across signature lengths.  This is to be expected if the neural embeddings provide a better measure of speech segment distances, since the increased number of permutations helps provide a more exact estimate of the embedding distance. We note that performance as measured in Table~\ref{tab:nn3:permsweep} has not plateaued in any of the Median Example metrics.  Further increasing the number of permutations may further improve these metrics, but this incurs a large cost in memory.

Figure~\ref{fig:Pat10} and Table~\ref{tab:nn3:bwsweep} show that, except for the cases with short signatures and few permutations, increasing beamwidth does not improve P@10 performance, while incurring significant cost.  To obtain higher precision systems, it is more important to use computational resources for increasing the number of permutations or using longer signatures.  However, as would be expected, the higher beamwidths help to significantly improve the FOM score, a metric concerned primarily with recall.

\subsection{Evaluation set performance}
Based on development results, we find that an operating point of 16 permutations, beamwidth of 2000, and signature length of 1024 is close to optimal, in terms of both performance and query speed, for both the baseline S-RAILS and S-RAILS+NAWE. We use these settings for final evaluation. For a qualitative view, Figure~\ref{fig:queries_and_top_hits} visualizes several queries and their top hits in the evaluation collection.  This visualization shows some expected properties.  For example, the two ``Massachusetts'' queries and their top hits are embedded close together.  Two of the false alarms for ``Massachusetts'' are the similar-sounding ``messages'' and ``math is just'', while the somewhat more distant ``math and science'' is (correctly) not retrieved.

Final evaluation performance is shown in Table~\ref{tab:eval}.  Besides the S-RAILS baseline, we also compare to RAILS~\cite{JanVan2012}, a DTW-based system that is optimized for speed using LSH to get approximate frame-level near neighbor matches. RAILS evaluation scores are reproduced from \cite{JanVan2012}. We find that our approach improves significantly over both RAILS and S-RAILS in terms of all performance metrics at this operating point.  Note that, based on Figure~\ref{fig:Pat10}, the improvements should hold at most operating points, including ones with much higher query speeds. The biggest gains from S-RAILS+NAWE are seen in the Median Example results, where there is a relative improvement over S-RAILS of more than 55\% across all measures. In terms of FOM and OTWV, we see relative improvements of over 40\% in the Best Example case. Although the baselines obtain good P@10, we still find large improvements in this measure as well, from 87.1\% to 95.1\%.

\section{Conclusion}
We have presented an approach to query-by-exmaple speech search using neural acoustic word embeddings, demonstrating the ability of these embedding models to improve over previous methods on a realistic task. The neural embeddings are learned from a very limited set of data; one interesting future direction is to study the limits of the approach as the amount of training data is varied, or to extend it to use no labeled data at all. Another interesting aspect of the approach is that the neural embeddings are learned from speech segments that have been pre-segmented at word boundaries, but they are then applied for embedding arbitrary segments that may or may not (and usually do not) correspond to words.  It is encouraging that this approach works despite the lack of non-word examples in the training data, and an interesting avenue for future work is to attempt to further improve performance by explicitly training on both word and non-word segments. Additional future directions include training a QbE system end-to-end and extending our model to operate at the level of multi-word phrases.

\vspace{-.05in}
\section{Acknowledgements}
This material is based upon work supported by the National Science Foundation under Grant No.~IIS-1433485 and by a Google faculty award.

\newpage\balance
\bibliographystyle{IEEEtran}
\bibliography{interspeech2017_refs}

\end{document}